\documentclass[twocolumn]{article}

\usepackage{setting}

\SetTitleLeft{}
\SetTitleRight{}
\SetHeaderAuthor{} 
\begin{document}

\begin{TitlePage}

%
%
%

\FullTitle{Robot and Overhead Crane Collaboration Scheme to Enhance Payload Manipulation}

%

\Authors{
    \textit{Antonio Rosales}\\
     VTT Technical Research Centre of Finland Ltd, Oulu, Finland

    \smallskip
    \textit{Alaa Abderrahim}\\
     VTT Technical Research Centre of Finland Ltd, Oulu, Finland

    \smallskip
    \textit{Markku Suomalainen}\\
     VTT Technical Research Centre of Finland Ltd, Oulu, Finland 
    
    \smallskip
    \textit{Mikael Haag}\\
    Konecranes, and
    
     \smallskip
    \textit{Tapio Heikkilä}\\
     VTT Technical Research Centre of Finland Ltd, Oulu, Finland 
}


\begin{Abstr}

This paper presents a scheme to enhance payload manipulation using a robot collaborating with an overhead crane. In the current industrial practice, when the crane's payload has to be accurately manipulated and located in a desired position, the task becomes laborious and risky since the operators have to guide the fine motions of the payload by hand.  In the proposed collaborative scheme, the crane lifts the payload while the robot's end-effector guides it toward the desired position. The only link between the robot and the crane is the interaction force produced during the guiding of the payload. Two admittance transfer functions are considered to accomplish harmless and smooth contact with the payload. The first is used in a position-based admittance control integrated with the robot. The second one adds compliance to the crane by processing the interaction force through the admittance transfer function to generate a crane's velocity command that makes the crane follow the payload. Then the robot's end-effector and the crane move collaboratively to guide the payload to the desired location. A method is presented to design the admittance controllers that accomplish a fluent robot-crane collaboration. Simulations and experiments validating the scheme potential are shown.
 
\end{Abstr}
%


\end{TitlePage}



\section{Introduction}
Overhead cranes are essential for lifting and moving weighty payloads in heavy manufacturing industries. When the payload needs to be located at a specific place or positioned at a desired pose, the crane's operator manually guides the payload either pushing or pulling to get the desired position. This manual guiding is mainly made with one hand while the other is used to operate the crane's control. Manual assistance requires skilled persons to be found and/or trained representing a time consumption not always welcome for the tight industrial production schedules \cite{HoffmanRAL2020}. Also, manual guiding might compromise the safety of the operators \cite{bey2022intelligent}, and if the guiding is not executed precisely, the payload might be damaged \cite{HoffmanRAL2021}.

Increasing the automation levels in overhead cranes is needed to ensure payload manipulation without risking the operator and the payload. Automation of payload manipulation has been carry on by the integration of robotic and mechatronic systems to cranes. One approach is cable-driven parallel robotic systems combined with current overhead crane technologies. This approach is presented in \cite{HoffmanRAL2020,HoffmanRAL2021,ONeillRAL21,ONeillRAL22} accomplishing fully automated insertion tasks only analyzing the cable tension forces. However, a fully automated solution misses human guidance and supervision capabilities. 

Another approach to automate overhead cranes is using Intelligent Assist Devices (IAD) \cite{kruger2009cooperation}. IAD are widely used in industrial applications to assist the operator in moving and lifting the payload \cite{colgate2008safety}, these devices transform the operator's forces and/or changes of payload's positions into crane commands. Considering the type of apparatus used to assist the crane, the IAD can be divided into two groups. One group employs handles or levers that map the force exerted on them to crane motion commands. The second group uses a robot arm to move the payload lifted by the crane.

Most of the IAD presented in the literature use handles/levers. In \cite{campeau2017articulated}, the pulling and pushing forces at the device's assistance are measured and analyzed but these forces are not used in the controller. In \cite{Campeau-LecoursTranMech16}, the authors integrate admittance force control to the approach in \cite{campeau2017articulated}, but no details about the design of the admittance controller are provided. The assistance device presented in \cite{WelchRAL22} uses admittance control including stability analysis, however, the accuracy of the payload position is compromised since the operator sets the desired position via his/her visual feedback. Another approach that fits in the IAD using handles/levers is the work in \cite{PengICCA09}. The authors used a tag held by the operator to sense three-dimensional motion and the sensed motion is used to command the crane. However, the method lacks the advantage of guiding the load directly since the operator indirectly guides the payload via the handled tag. 

A few works focus on using a robot as an IAD, and most of them are based on constraint motion i.e. only position control is used for controlling the interaction between robot and payload, see \cite{ambrosinoISARC2022,ambrosino2024}. In \cite{SchubertMed19}, a robot operated with a joystick is the assistant device. Force feedback between the assistance device (robot) and the joystick is considered, but the robot's and the crane's interaction is based on constraint motion. Using constraint motion to execute interaction tasks is not recommended since contact forces can increase and saturate the robot's actuators or the object in contact can be damaged \cite{SicilianoBook09}. In \cite{AraiISARC1988}, the IAD is a robot with a flexible link to add compliance and move the crane's payload smoothly. The signal of a strain gauge mounted at the flexible link is used to sense the interaction between the robot and the payload. The main drawback of the approach is the flexible link since oscillations may occur and changing the stiffness requires a physical modification of the robot. Also, there are patents addressing the manipulation of heavy loads considering a crane collaborating with a robot, and using force measurements \cite{Patent1,Patent2,Patent3}, however the patents omit details of the admittance controller used for mapping force to velocity.


This paper presents a novel robot and crane collaborative scheme to manipulate payloads integrating for first-time compliance into the crane via admittance control. The scheme considers a robot guiding a payload lifted by the crane. The crane and robot's end-effector move collaboratively to drive the payload at a desired velocity. The collaboration is based on the interaction force between the robot's end-effector and the payload. The robot and the crane are integrated with admittance controllers to accomplish a soft and safe interaction. The interaction force is measured and used to implement a position-based admittance controller in the robot. On the crane side, the measured force is sent to the crane through an admittance controller to generate a velocity command. The design and stability analysis of the admittance controllers are presented. The functionality of the scheme is validated via simulations and experiments.

Compared with the IAD approaches using handlers/levers presented in \cite{campeau2017articulated,Campeau-LecoursTranMech16,WelchRAL22,PengICCA09}, the proposed scheme is harmless for the operator since the robot interacts directly with the payload, and the operator can supervise the manipulation or command the robot using a joystick from a risk-free place.  Also, in comparison with the fully automated methods presented in \cite{HoffmanRAL2020,HoffmanRAL2021,ONeillRAL21,ONeillRAL22}, the proposed scheme does not remove the valuable skills and experience of the operator since he/she can still supervise or manipulate the robot. Considering the IAD using a robot presented in \cite{SchubertMed19} and \cite{AraiISARC1988}, this approach includes compliance in the robot and crane via admittance control offering an accessible way to modify stiffness and damping. Furthermore, the paper presents the design and stability analysis of the admittance controls implemented on the robot and the crane.


The structure of the paper is the next one. Section II contains the description of the proposed robot crane collaboration scheme. The design and analysis of the scheme are presented in Section III. Section IV includes the simulation and experiments, and Section V has the conclusions.

\section{Problem statement}
\begin{figure}
	\centering
	\includegraphics[width=9cm]{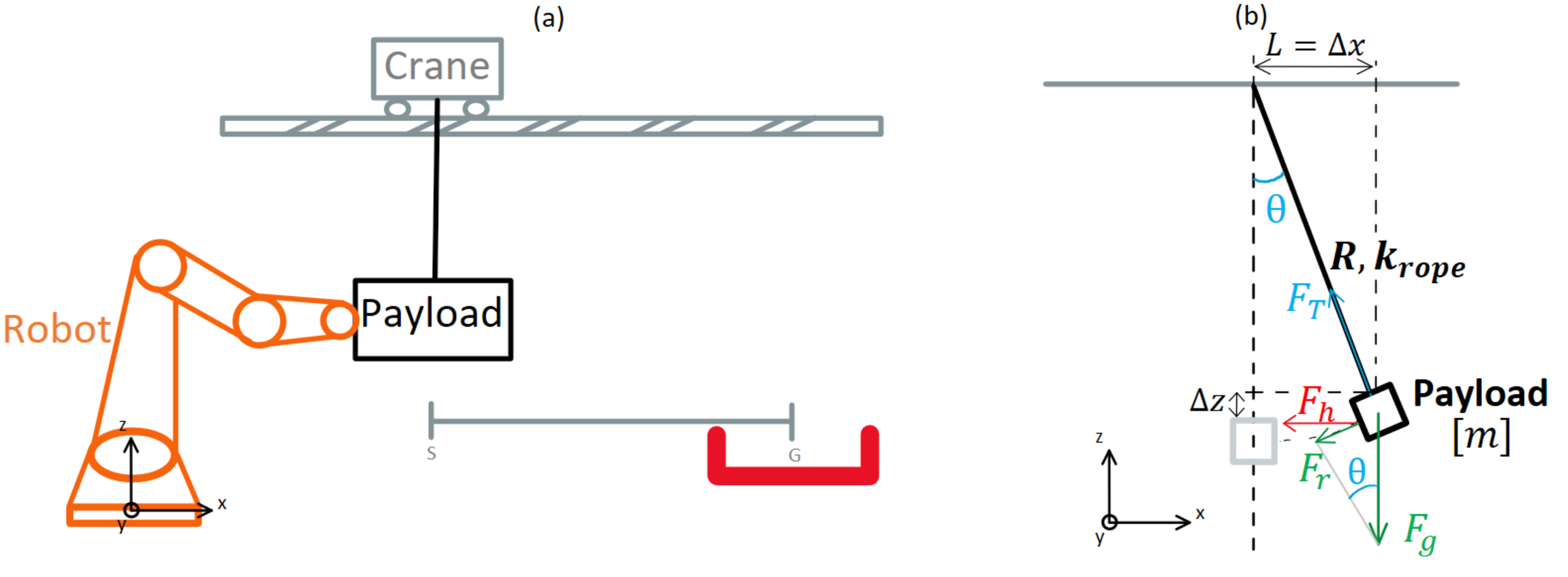}
	\caption{ (a) Sketch of the collaborative task robot and crane. (b) Pendulum model and contact forces}
	\label{fig:task1}
\end{figure}

Consider a robot in contact with a payload lifted by a crane, see Fig. \ref{fig:task1}-(a). The goal is to use the robot to guide the payload from a starting point $S$ to a final point $G$, while the crane lifts the payload, i.e. the robot collaborates with the crane to accurately locate the payload in a target position. The robot and crane collaboration is based on the contact force exerted on the payload by the robot's end-effector. The payload can be guided in three directions $(x,y,z)$, and the movement in each direction is decentralized, e.g. no direct communication between them. The decentralized guiding is easy to accomplish by controlling the robot and crane in Cartesian space. In the paper, we focus on the horizontal direction $x$ but the approach can be easily extended to the motion on the plane $XZ$ or in the 3D space $XYZ$.

Consider the robot's end-effector exerts a force on the payload along horizontal direction $x$ producing an angle $\theta$ measured from the vertical position, see Fig \ref{fig:task1}-(b). The displacement on $x$ direction can be analyzed using a pendulum model. The payload's mass $m$ is the pendulum's mass, $R$ is the length of a mass-less rope and $k_{rope}$ is the rope constant. $\theta$ represents the sway angle, $L$ is the horizontal displacement, $F_g$ is the gravity force , $F_T$ is tension along the rope, $F_r$ is the pendulum's restoring force ,and $F_h$ is the horizontal force applied at the end-effector. Considering $F_g=mg$, with $g$ the earth's gravity, $F_R=-F_g\sin{\theta}$ , $F_T=k_{rope}\Delta z$ ,and $\theta=\arcsin{(\frac{L}{R})}$, the horizontal force $F_{h}$ is computed as follows,
\begin{equation}\label{eq:Fh}
	F_{h}=F_R\cos{\theta}+F_T\sin{\theta}=\sin{\theta}(-F_{g}\cos{\theta}+F_T)
\end{equation}
The force $F_{h}$ can be studied as an elastic interaction force $F$ between the robot and the crane. Replacing $\sin{\theta}$ with $\frac{L}{R}$, one gets $ F_{h}=\frac{L}{R}(-F_{g}\cos{\theta}+F_T)$, and the model of the interaction force is,
\begin{equation}\label{eq:Elasmodel}
	F=K_{e}\Delta x
\end{equation}
where $\Delta x=L$ is the difference between the positions of the end-effector/payload and the crane along the $x$-axis (see Fig \ref{fig:task1}-(b)), and $K_e=(F_T-mg\cos{\theta})/R$ is the environment's stiffness. 
Note that the displacement $\Delta x$ is directly related to the angle $\theta$. When $\theta=0$, $\Delta x=0$ since the crane and end-effector are in the same position. The angle $\theta \neq 0$, when there is a difference in the position of the end-effector/payload compared with the crane, caused by the robot pushing the payload along the $x$~axis. Also, one can see that the environment's stiffness $K_e$ depends on the payload mass $m$ and the rope length $R$, the heavier the mass the stiffer the environment. 

\begin{figure}
	\centering
	\includegraphics[width=9cm]{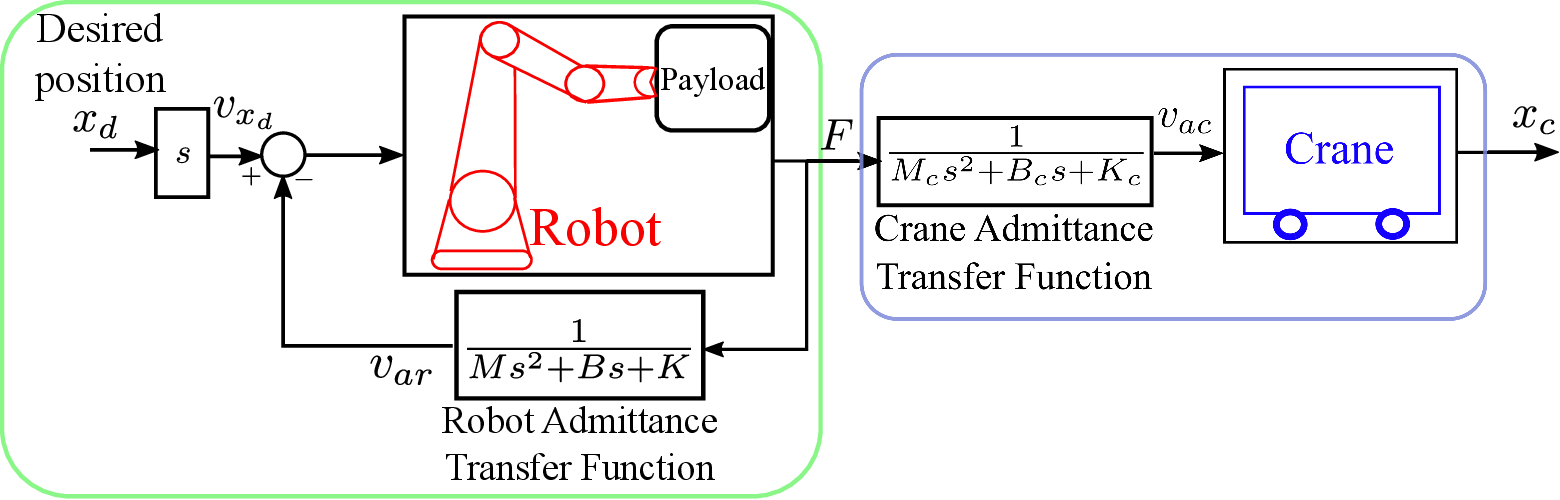}
	\caption{Block diagram of the collaborative scheme}
	\label{fig:bdscheme}
\end{figure}

A collaborative scheme with two admittance controllers is proposed to achieve a smooth robot-crane collaboration when the payload is manipulated. The block diagram of the scheme is presented in Fig. \ref{fig:bdscheme}. On the robot side, a position-based admittance control, see \cite{vukobratovic2009dynamics}, ensures harmless contact with the payload while a desired position $x_d$ or velocity $v_{x_d}$ is reached. The admittance transfer function integrated into the robot's control loop makes the robot behave like a mass-spring-damper system with parameters $M_r$, $B_r$, and $K_r$ to be selected. On the crane side, the admittance transfer function with parameters $M_c$, $B_c$, and $K_c$ transforms the interaction force $F$ into velocity commands $v_{ac}$ needed to track the velocity set by the robot. The transformed velocity $v_{ac}$ is characterized by the mass-spring-damping response defined by $M_c$, $B_c$, and $K_c$. The interaction force $F$ is the only signal connecting the robot with the crane.

The next section presents how the admittance parameters $M_r$, $B_r$, $K_r$, $M_c$, $B_c$, and $K_c$ should be selected to accomplish payload manipulation via robot-crane collaboration.

\section{Robot-Crane Collaboration Scheme}
This section describes the details of the proposed robot-crane collaboration scheme. Firstly, the admittance controllers implemented on the robot and crane are presented. Then, a procedure for designing the robot's and crane's admittance parameters is given. The last part of the section shows a method to verify the stability of the whole system i.e. robot admittance controller working together with the crane admittance controller.

\subsection{Robot and Crane Admittance Controllers}
The admittance transfer functions on the robot and crane sides are integrated into a closed-loop and an open-loop system, respectively, see Fig. \ref{fig:bdscheme}. The selection of the robot admittance transfer function parameters must consider the closed-loop stability including the robot's dynamics. On the other hand, the crane admittance transfer function defines an open-loop system together with the crane dynamics, and the selection of the admittance parameters is mainly to shape the velocity command $v_{ac}$ from the received force $F$. A closed-loop velocity control between the crane and its admittance seems a natural option but analyzing an open-loop system is better from a practical perspective since the closed hardware architecture of the cranes rarely provides velocity measurements. 

The robot and crane admittance transfer functions of the scheme in Fig. \ref{fig:bdscheme} can be represented as a second-order transfer function 
\begin{equation}\label{eq:adrobcra}
	\frac{V_{a_{i}}(s)}{F(s)} =\frac{1}{M_{i}s^2+B_{i}s+K_{i}}=\frac{\omega_{n_{i}}^2/M_{i}}{s^2 + 2\zeta_{i}\omega_{n_{i}}s + \omega_{n_{i}}^2}
\end{equation}
where $s$ is the Laplace variable, $\omega_{n_{i}}$ is the natural frequency, and $\zeta_{i} $ is the damping coefficient.  The subscript $i$ refers to the coefficients of the robot admittance transfer function when $i=r$, and to the coefficients of the crane admittance transfer function when $i=c$. Then, the robot's admittance parameters are $M_r$, $B_r$, and $K_r$, and the crane's admittance parameters  are $M_c$, $B_c$, and $K_c$. 

From eq. (\ref{eq:adrobcra}), the natural frequency $\omega_{n_{i}}$, and the damping coefficient $\zeta_{i} $ can be written in terms of the admittance parameters $M_i$, $B_i$, and $K_i$ as follows, $\omega_{n_r} = \sqrt{\frac{K_r}{M_r}}$, $\zeta_r = \frac{B_r}{2\sqrt{M_rK_r}}$, $\omega_{n_c} = \sqrt{\frac{K_c}{M_c}}$, and $\zeta_c = \frac{B_c}{2\sqrt{M_cK_c}}$. Thus, the time response of the robot and crane admittance controllers is characterized by the values of $\omega_{n_r}$ and $\zeta_r$, and $\omega_{n_c}$ and $\zeta_c$, respectively. Therefore, the robot's admittance parameters $M_r$, $B_r$, and $K_r$ that provide a desired time response can be computed using eq. (\ref{eq:adrobcra}). Also, the crane's admittance parameters $M_c$, $B_c$, and $K_c$ that give a desired time response can be computed using eq. (\ref{eq:adrobcra}).

\subsection{Robot's admittance control design}
\begin{figure}
	\centering
	\includegraphics[width=6cm]{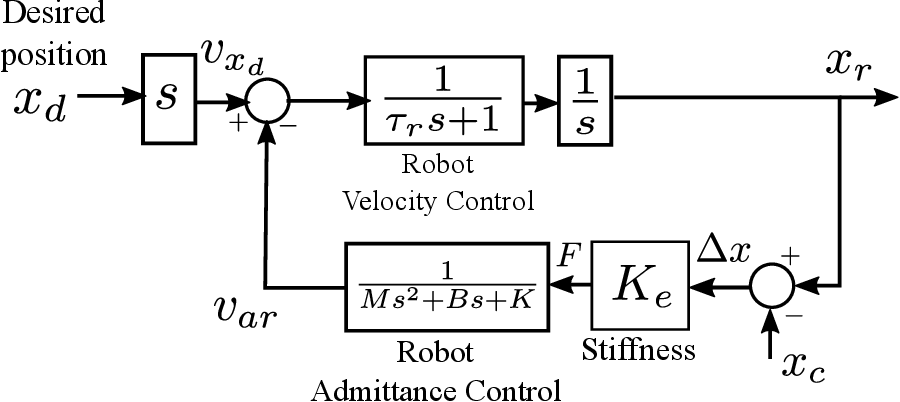}
	\caption{Block diagram of position-based admittance control}
	\label{fig:bdadrob}
\end{figure}

Consider the position-based admittance control in the block diagram in Fig. \ref{fig:bdadrob}, see \cite{vukobratovic2009dynamics}. The robot's dynamics are studied using a velocity controller transfer function with time constant $\tau_r$. The block $K_e$ is the stiffness of the environment used to compute the force $F$ in eq. (\ref{eq:Elasmodel}). From Fig. \ref{fig:bdadrob}, the transfer function from the input $x_d$ to the output $x_r$ is 
\begin{equation}\label{eq:robadtf}
	\frac{X_r(s)}{X_d(s)}=\frac{s(M_rs^2+B_rs+K_r)}{c_{1}s^{4}+c_2s^3+c_3s^{2}+c_{4}s+c_5}
\end{equation}
where $c_1=\tau_{r}M_r$, $c_2=\tau_{r}B_r+M_r$, $c_3=\tau_{r}K_r+B_r$, $c_4=\tau_{r}K_{e}+K$, $c_{5}=K_e$, and $s$ is the Laplace variable. The denominator in eq. (\ref{eq:robadtf}) is the characteristic equation of the system \cite{bishop2011modern}, and it provides information about the system's stability. When all the roots of $c_{1}s^{4}+c_2s^3+c_3s^{2}+c_{4}s+c_5$ have the real part negative, one can conclude the system is stable.

The robot's admittance control parameters $M_r$, $B_r$, and $K_r$ are selected using the second-order system representation in eq. (\ref{eq:adrobcra}), and the transfer function in eq. (\ref{eq:robadtf}) is employed to verify stability. 

Selecting a large value of damping coefficient $\zeta$ is a common approach to achieve a critical damping response avoiding oscillations during contact, see \cite{vukobratovic2009dynamics}. Then, a damping factor of $\zeta_r=1$ is chosen to have a response with critical damping, and from the equation $\zeta_r = \frac{B_r}{2\sqrt{M_rK_r}}$, the mass $M_r$, the stiffness $K_r$, and the damping $B_r$ are linked by the equation,
\begin{equation}\label{eq:bzeta1}
	B_r = 2\sqrt{M_r \cdot K_r},
\end{equation}

Using eq. (\ref{eq:bzeta1}), the procedure to select the robot's admittance control parameters starts by choosing the value of the mass $M_r$, and a stiffness value $K_r$ bigger than the environment stiffness $K_e$ to have a rigid robot capable of moving the payload. Then, the damping $B$ that gives a critical damping response is selected using eq. (\ref{eq:bzeta1}). 

The stability of the selected parameters should be tested using the characteristic equation in (\ref{eq:robadtf}). A useful way to check stability is observing the location of the roots of the characteristic equation in (\ref{eq:robadtf}) when the parameters $M_r$, $B_r$, and $K_r$ change. For example, one can know how big the value of $K_r$ has to be selected to preserve stability. The next numerical example shows how the root's location can be obtained, and how the system stability can be verified.

\subsubsection{Numerical example} 
\begin{figure}
	\centering
	\includegraphics[width=7cm]{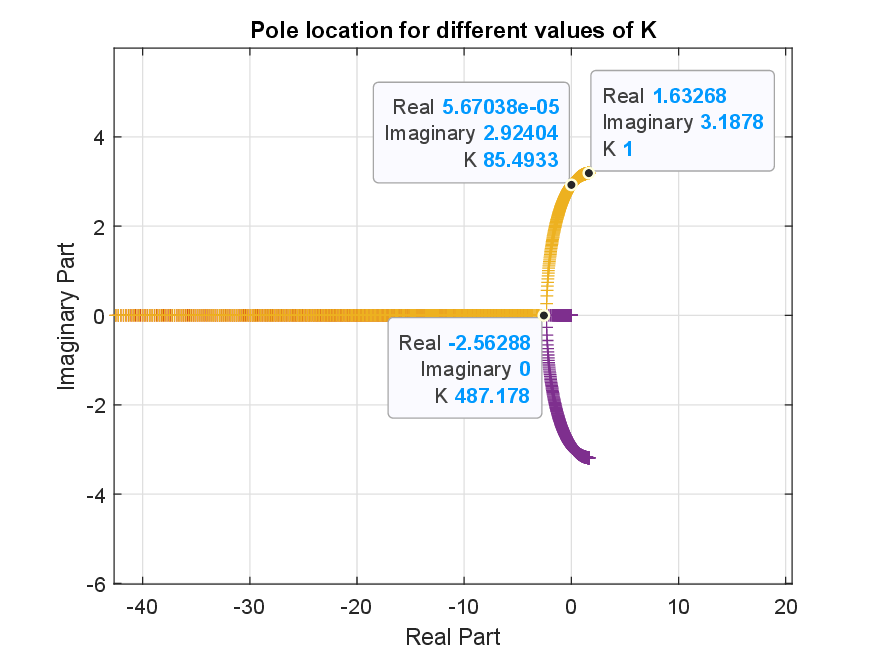}
	\caption{Location of roots of the characteristic equation in eq. (\ref{eq:robadtf}) for different values of $K_r$ and $B_r$ with $M_r=10$}
	\label{fig:Polesrob}
\end{figure}

Considering the time constant $\tau_r=0.02$, the environment's stiffness $K_e=500$ (equivalent to a pendulum of mass $m=100$~[kg], rope length $R=2$~[m], and $g=9.81$~[m/s$^2$], see Fig. \ref{fig:FhvsL}), and $M_r=10$. Once the mass is fixed as $M_r=10$, one can compute a set of damping values $B_r$ from a set of $K_r$ values using eq. (\ref{eq:bzeta1}). The set of $B_r$ and $K_r$ values form a set of characteristic equations with roots located at different places of the imaginary and real axes. For example, for a set of values of $K_r=[1,2,3,... , 100000]$, a set of values of $B_r$ is obtained, and the roots location for the corresponding set of characteristic equations is presented in Fig. \ref{fig:Polesrob}. Three values of $K_r$ are marked in in Fig. \ref{fig:Polesrob}.One value corresponds to $K_r=1$ with roots located on the right side of the complex plane. The second value $K_r=85.49$ is a critical value located on the imaginary axis, and the third value $K_r=487.178$ corresponds to roots on the real axis. Therefore, one must avoid choosing $K_r<85.49$ since the roots are located on the right side and instability is expected. On the other hand, choosing $K_r \geq 487.178$ produces a non-oscillatory response, and an oscillatory behavior is expected when $85.49 < K_r < 487.178$.

\subsection{Crane's admittance control design}
A similar approach can be followed to select the crane's admittance parameters $M_c$, $B_c$, and $K_c$. In the crane's case, one should consider critical damping via $B_c = 2\sqrt{M_c \cdot K_c}$, and the crane dynamics using the transfer function,
\begin{equation*}
	\frac{X_c(s)}{V_{ac}(s)} =\frac{1}{s(\tau_c s+1)}
\end{equation*}
where $\tau_c$ is the time constant corresponding to the crane's velocity control, $X_c(s)$ and $V_{ac}(s)$ are the crane's position $x_c$ and velocity $v_{ac}$ in the Laplace domain, respectively. 

The transfer function from the force $F(s)$ to the crane's position $X_c(s)$ is,
\begin{equation}\label{eq:craneadtf}
	\frac{X_c(s)}{F(s)}=\frac{1}{s(\tau_c s+1)(M_c s^2+B_c s+ K_c)},
\end{equation}
obtained via the cascade connection of the crane's admittance transfer function in eq. (\ref{eq:adrobcra}) and the transfer function of the crane dynamics $1/(s(\tau_c s+1))$, see Fig. \ref{fig:bdscheme}.

The mass $M_c$ and stiffness $K_c$ should be selected considering the robot's admittance parameters in the following way. The virtual mass $M_c$  should be lighter than $M_r$ to ensure the robot can push the payload. The stiffness $K_c$ should be smaller than $K_r$ to have a complaint crane that moves after the robot pushes the payload. The stability of the crane's admittance can be verified by analyzing the roots of the characteristic equation of transfer function in eq. (\ref{eq:craneadtf}). The next example shows how to select the crane's admittance parameters and verify stability. 

\subsubsection{Numerical example} 
Consider the time constant $\tau_c=0.1$, the environment's stiffness $K_e=500$, and the robot's admittance parameters from the previous numeric example $M_r=10$, $B_r=283$, and $K_r=2000$. Then, the selection of the crane's admittance parameters is the next. The mass $M_c=1$ and the stiffness $K_c=1000$ are selected smaller than $M_r$ and $K_r$, respectively. The damping $B_c$ is computed as $B_c=2\sqrt{1*1000}=64$. Using the selected parameters $M_c=1$, $B_c=64$, and $K_c=1000$ in the characteristic equation of eq. \ref{eq:craneadtf}, the roots are  $0,-32.5536,-30.6857,-10.0107$, and stability in the crane's admittance controller is expected.

\subsection{Stability analysis of the collaborative scheme}
\begin{figure}
	\centering
	\includegraphics[width=6cm]{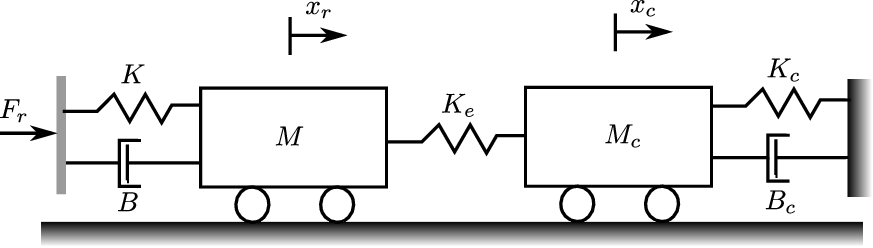}
	\caption{Equivalent two-mass spring damper system}
	\label{fig:Twomass}
\end{figure}

The proposed scheme in Fig. \ref{fig:bdscheme} can be analyzed using two mass-spring-damper models. One model is the equivalent mass-spring-damper system of the robot under admittance control, and the second model is the admittance of the crane. The two equivalent models are connected through the stiffness of the environment $K_e$, see Figure \ref{fig:Twomass}. 

The dynamics of the system presented in Figure \ref{fig:Twomass} are defined by the next equations,
\begin{eqnarray}
	M_r\ddot{x}_r+B_r\dot{x}_r+K_rx_r&=&F_r+F \label{eq:msd1}\\
	M_c\ddot{x}_c+B_c\dot{x_c}+K_{c}x_{c}&=&F \label{eq:msd2}
\end{eqnarray}
where $x_r$, $M_r$, $B_r$, and $K_r$ are the robot's position, the mass, the damping, and the stiffness of the robot's admittance, respectively. The force produced by the robot's actuators is $F_r$, and $F$ is the interaction force defined by the elastic model in eq. (\ref{eq:Elasmodel}) with environment stiffness $K_e$, and $\Delta=x_r-x_c$. The crane's position and its admittance parameters are $x_c$, $M_c$, $B_c$, and $K_c$, respectively.

Considering initial conditions equal to zero, and applying the Laplace transform to equations (\ref{eq:msd1}) and (\ref{eq:msd2}), the transfer function from the force $F_r$ to position $x_c$ is the next one
\begin{equation}\label{eq:tf}
	\frac{X_{c}(s)}{F_{r}(s)}=\frac{K_e}{s^4+a_1s^3+a_2s^2+a_3s+a_4}
\end{equation}
where $s$ is the Laplace variable, and  $X_{c}(s)$ and $F_{r}(s)$ are the crane's position $x_c$ and the robot's force $F_r$ in Laplace domain, respectively. The denomitator's coefficients are $a_1=\frac{M_rB_c+B_rM_c}{M_rM_c}$, $a_2=\frac{M_rK_c+M_rK_e+B_rB_c+K_rM_c-K_eM_c}{M_rM_c}$, $a_3=\frac{B_rK_c+B_rK_e+K_rB_c-K_eB_c}{M_rM_c}$, and $a_4=\frac{K_rK_c+K_rK_e-KeKc}{M_rM_c}$.

The transfer function in eq. (\ref{eq:tf}) is an input-output model of the collaborative scheme in Fig. \ref{fig:bdscheme}. Eq. (\ref{eq:tf}) describes how the crane's position $x_c$ responds to the force $F_r$ produced by the robot and transmitted to the payload via the interaction force $F$. The Routh-Hurtwitz stability criterion \cite{bishop2011modern} can be used to verify the admittance parameters that ensure the stability of the whole scheme.

Considering the Routh-Hurtwitz stability criterion, and using the denominator coefficients of eq. (\ref{eq:tf}), the admittance parameters of the robot and the crane that ensure stability have to satisfy the next inequalities (see \cite{bishop2011modern}),
\begin{equation}
	a_1>0; \quad a_3>0; \quad a_4>0 \quad a_1a_2a_3>a_3^2+a_1^2a_4 \label{eq:sta1}
\end{equation}

Finding an analytical solution for the inequalities in eq. (\ref{eq:sta1}) is not straightforward, however, they can be implemented using a numeric computing software like Matlab, and the stability can be verified for selected admittance parameters. The next example shows how the inequalities (\ref{eq:sta1}) are used to verify stability.

\subsubsection{Numerical example}
Considering the environment's stiffness $K_e=500$, and the robot and crane admittance parameters from the previous examples $M_r=10$, $B_r=283$, $K_r=2000$, $M_c=1$, $B_c=64$, $K_c=1000$. The coefficients $a_1=92.3$, $a_2=3461.2$, $a_3=52050$, and $a_4=250000$ of the transfer function in eq. (\ref{eq:tf}) are computed. Then, the inequalities in eq. (\ref{eq:sta1}) hold since $a_1>0$ $a_3>0$, $a_4>0$, and $a_1a_2a_3>a_3^2+a_1^2a_4$ ($1.6\times 10^{10}>4.8\times 10^{9}$). Therefore, the stability of the whole collaborative scheme is concluded.

\section{Numerical Simulations and Robot Experiment}
This section presents the validation of the proposed collaborative scheme via numeral simulations and experiments using a lightweight robot to push the payload and an industrial robot with a pendulum attached to its end effector to emulate the crane and payload.

\subsection{Numerical simulations}
\begin{figure}[h] 
	\centering
	\includegraphics[width=8cm]{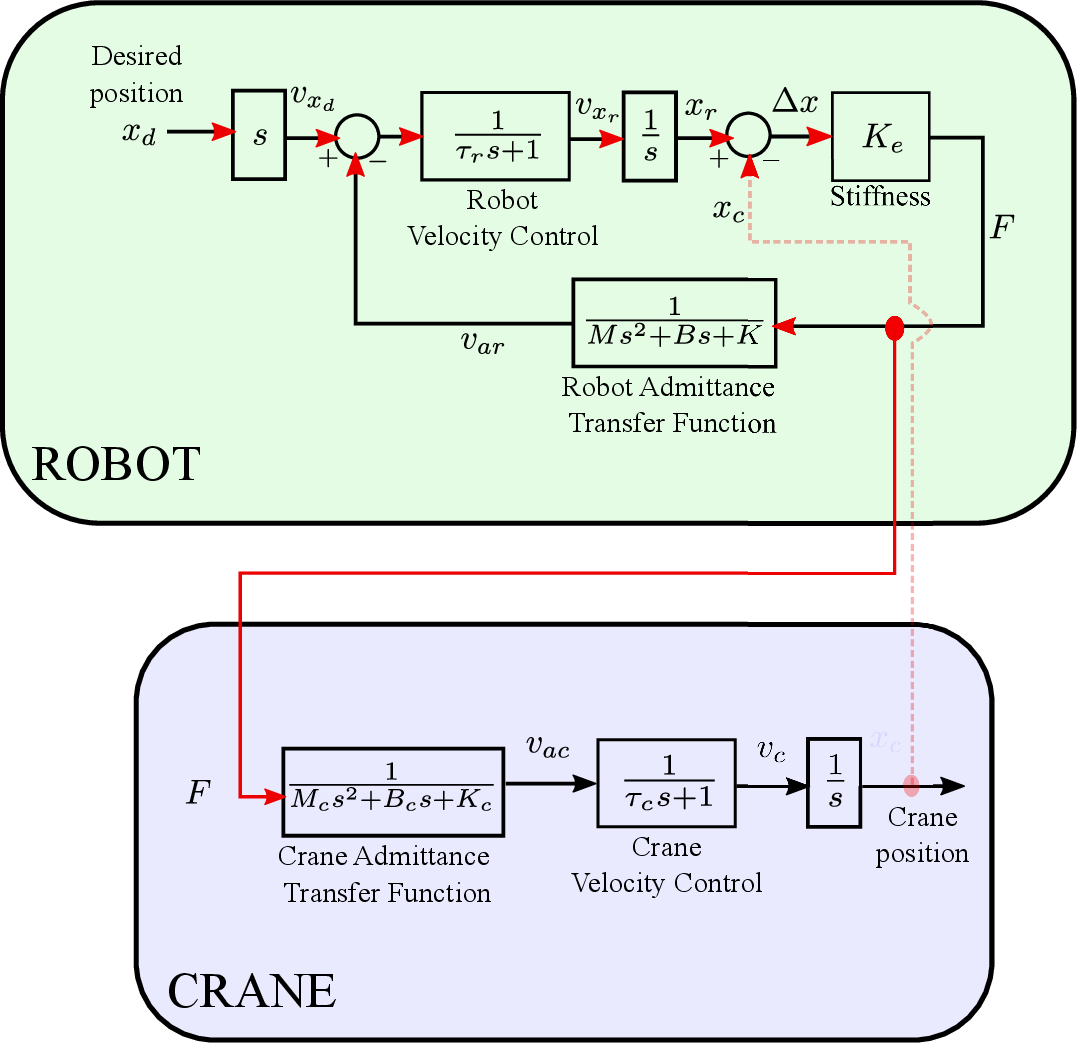}
	\caption{Simulation block diagram of the collaborative scheme}
	\label{fig:bdcoll}
\end{figure}

To validate the functionality of the proposed collaborative scheme in Fig. \ref{fig:bdscheme}, the scheme is programmed in Simulink following the block diagram in Fig. \ref{fig:bdcoll}. The robot's dynamics is simulated as the velocity control transfer function with time constant $\tau_r$, and an integrator to get the robot's position $x_r$. The crane's dynamics is simulated using the velocity control transfer function with time constant  $\tau_r$, and an integrator to get the crane's position $x_c$. Constant $K_e$ is the environment's stiffness used to compute the interaction force $F$ from $\Delta x=x_r-x_c$. The blocks robot and crane admittance transfer functions generate the velocities $v_{ar}$ and $v_{ac}$, respectively. Note that the crane's position $x_c$ is connected with the robot's loop to compute $F$ for simulation purposes. In practice, the interaction force $F$ is the only signal sent to the crane loop. 

\begin{figure}
	\centering
	\includegraphics[width=7cm]{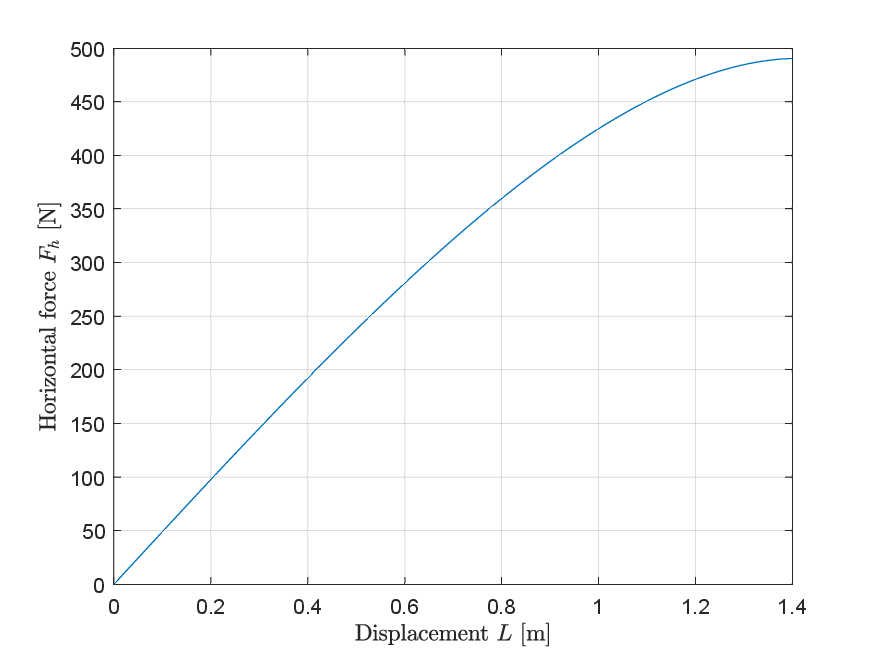}
	\caption{Plot horizontal force $F_h$ in eq. (\ref{eq:Fh}) versus displacement $L$ considering $m=100$~[kg], $R=2$~[m], and $g=9.81$~[m/s$^2$]}
	\label{fig:FhvsL}
\end{figure}

The simulation is performed using the following parameters. $\tau_r=0.02$ and $\tau_c=0.1$ are the time constants of the transfer functions representing the robot and crane velocity controllers, respectively. Considering a payload of mass $m=100$~[kg] with a rope's length $R=2$~[m], and the gravity $g=9.81$~[m/s$^2$], the stiffness $K_e$ equivalent to those parameters is estimated via eq. (\ref{eq:Fh}). Using a set of values of $L$ to get a set of values of $F_h$, a plot $L$ vs $F_h$ is obtained, and the value of $K_e=500$~[N/m] is the slope of the plot's linear part, see Figure \ref{fig:FhvsL}.

The admittance parameters for the robot and crane are $M=10$~[kg], $B=2000$~[Ns/m], $K=60000$~[N/m], $M_c=1$~[kg], $B_c=500$~[Ns/m], and $K_c=1000$~[N/m]. One can verify that these admittance parameters satisfy the stability conditions (\ref{eq:sta1}).

The simulation lasts 20 seconds with a fixed sample time of 4 milliseconds using Euler solver. The robot's velocity reference $v_{x_d}$ is a trapezoidal velocity profile with a maximum velocity of approximately 0.1~[m/s].

\begin{figure}
	\centering
	\includegraphics[width=9cm]{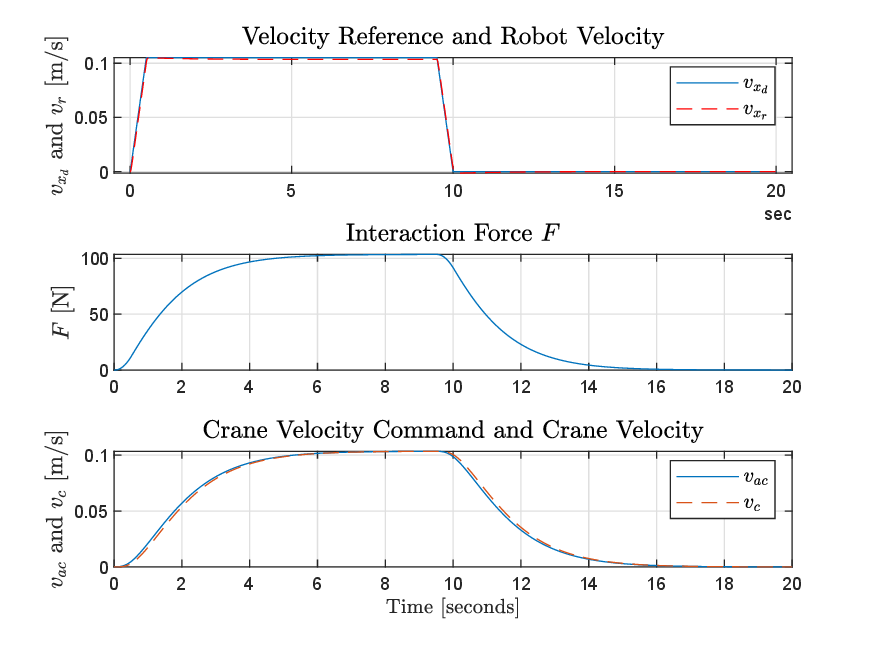}
	\caption{Simulation of the collaborative scheme with a maximum velocity 0.1~[m/s], $\tau_r=0.02$ and $\tau_c=0.1$. Top: Robot velocity reference $v_{x_d}$ and robot velocity $v_{x_r}$. Middle: contact force $F$. Bottom: crane velocity command $v_{ac}$ and crane velocity $v_c$  }
	\label{fig:Psimpos}
\end{figure}

The robot velocity reference $v_{x_d}$, robot velocity $v_{x_r}$, contact force $F$, crane velocity command $v_{ac}$, and crane velocity $v_c$ obtained from the numerical simulation are presented in Fig. \ref{fig:Psimpos}. The upper plot shows that the robot velocity $v_{x_r}$ follows the velocity reference $v_{x_d}$ exerting a force $F$ of maximum 100~[N] on the payload. From the contact force $F$, the crane admittance transfer function produces a crane velocity command $v_{ac}$ with a maximum velocity of 0.1~[m/s]. The crane velocity $v_c$ follows the command $v_{ac}$. 

From the numerical simulation results presented in Fig. \ref{fig:Psimpos}, one can conclude that the collaboration robot and crane is achieved since the crane moves according to the interaction force $F$ generated by the robot pushing the payload.

\begin{figure}
	\centering
	\includegraphics[width=8cm]{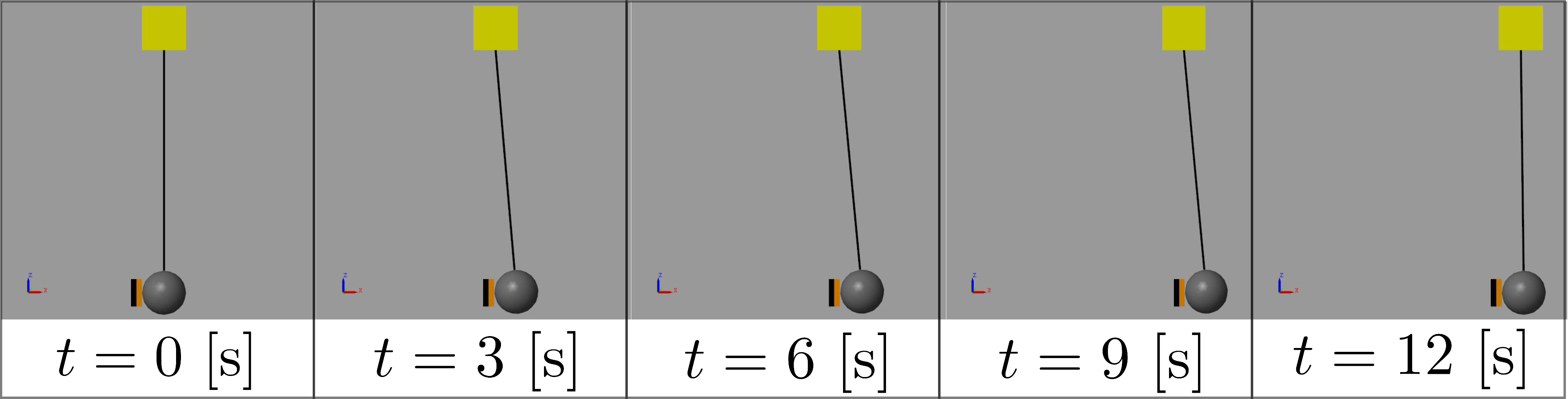}
	\caption{Frames of the 3D animation from Simscape. The robot end-effector, the rope, the payload, and the crane's trolley are represented by the orange/black rectangular brick, the black straight line, the grey sphere, and the yellow square brick, respectively.}
	\label{fig:framesim}
\end{figure}

Using Simscape, a 3D animation of the collaboration robot and crane is built inside the Simulink simulation. Fig. \ref{fig:framesim} presents the animation including the robot end-effector, the rope, the payload, and the crane's trolley represented by the orange/black rectangular brick, the black straight line, the grey sphere, and the yellow square brick, respectively.

The frames in Fig. \ref{fig:framesim} show the displacement of the robot's end-effector in contact with the payload, and the crane, when the trapezoidal profile presented in Fig. \ref{fig:Psimpos} is applied to the robot. One can observe how the robot moves the payload producing a sway angle different than zero, and a position deviation with respect to the crane. Then the crane moves after the payload until they reach the final position. Therefore, the 3D animation confirms that the robot and the crane manipulate the payload collaboratively.

\subsubsection{Collaborative scheme vs velocity control}
\begin{figure}
	\centering
	\includegraphics[width=9cm]{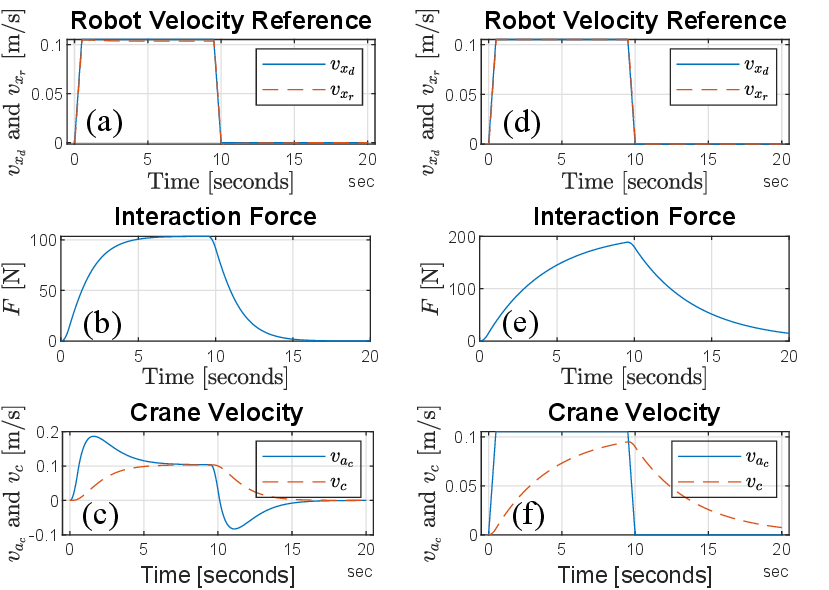}
	\caption{Simulation of the collaborative scheme and velocity control with a maximum velocity 0.1~[m/s], $\tau_r=0.02$ and $\tau_c=4$. Top: Robot velocity reference $v_{x_d}$ and robot velocity $v_{x_r}$. Middle: contact force $F$. Bottom: crane velocity command $v_{ac}$ and crane velocity $v_c$  }
	\label{fig:Psimposcom}
\end{figure}

A comparison of the collaborative scheme with only velocity commands in the crane and the robot is made to illustrate how pure velocity control might not be the best option for a collaboration robot-crane. Consider the previously used velocity time constant $\tau_r=0.02$ for the robot, and a velocity time constant $\tau_c=4$ slower than the previous one. The velocity control is tested by sending the same command $v_{x_d}$ (used previously) to the robot and the crane omitting interaction forces and admittance control. Plots of the collaborative scheme and the velocity control simulation are presented in Fig. \ref{fig:Psimposcom}-(a-c), and Fig. \ref{fig:Psimposcom}-(d-f), respectively. The interaction force $F$ and the crane's velocity $v_c$ are the main differences between the two approaches, see Fig. \ref{fig:Psimposcom}-(b-c) and \ref{fig:Psimposcom}-(e-f). 

From Fig. \ref{fig:Psimposcom}-(b) and Fig. \ref{fig:Psimposcom}-(e), one can observe that the velocity control generates an interaction force $F$ bigger than the proposed collaborative scheme. Comparing crane velocity $v_c$ in  Fig. \ref{fig:Psimposcom}-(c) and Fig. \ref{fig:Psimposcom}-(f), the presented collaborative scheme makes the crane move faster and follows the interaction force despite the slow dynamics of the crane. One can observe that the crane admittance control shapes the velocity command $v_{ac}$ to generate a crane velocity $v_c$ that follows the interaction force.

\subsection{Robot Experiment}
\begin{figure}
	\centering
	\includegraphics[width=7cm]{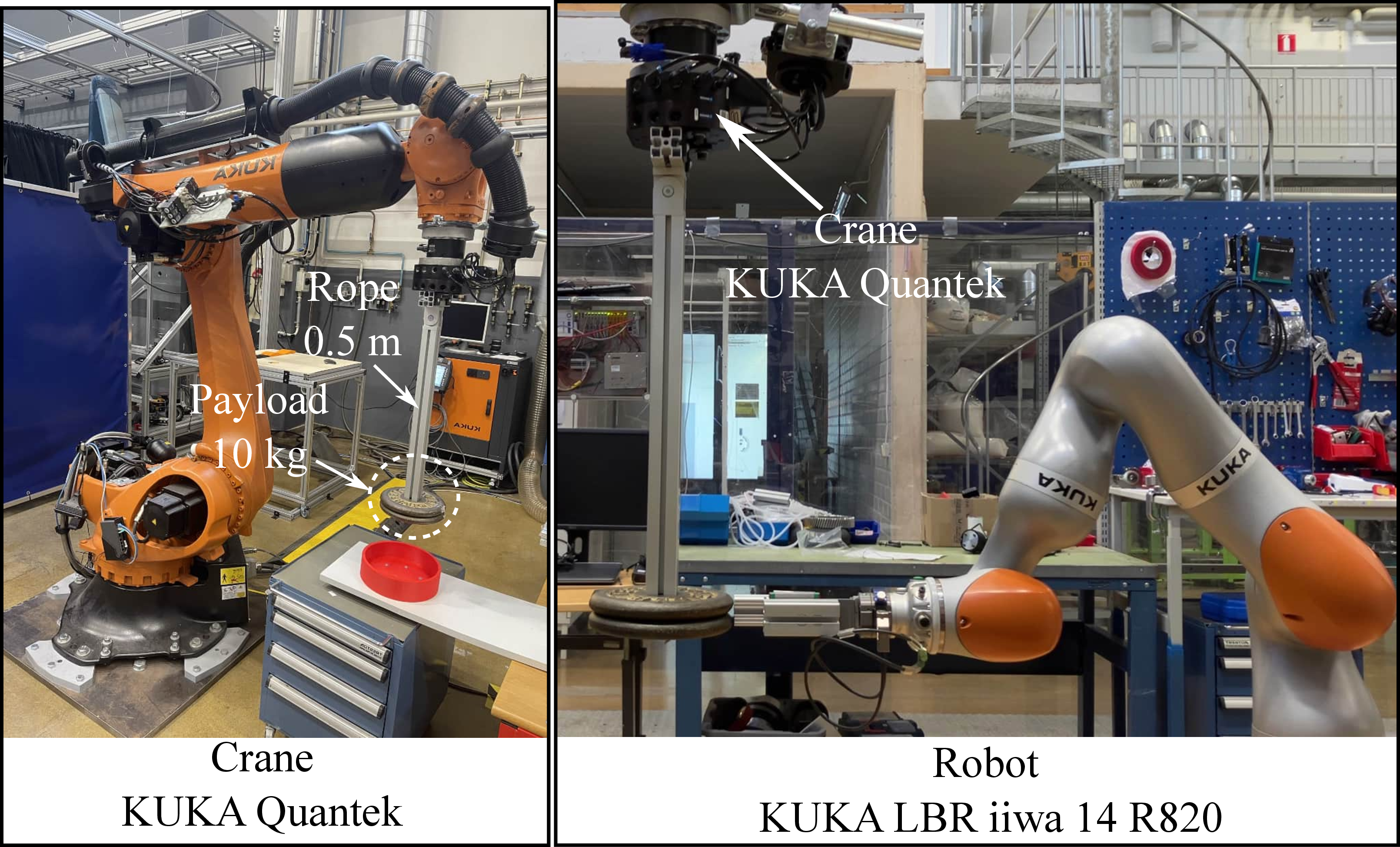}
	\caption{Experimental setup: KUKA KR 210 R2700 with attached pendulum emulates the crane, and KUKA LBR iiwa 14 R820 is the robot}
	\label{fig:expsetup}
\end{figure}

The proposed collaborative scheme is tested using the experimental setup in Fig. \ref{fig:expsetup}. The crane and payload are emulated by attaching a pendulum on an industrial robot KUKA KR 210 R2700 (Quantec). The attached mass and the length of the rope are $m=10$~[kg] and $R=0.5$~[m], respectively. The robot used to move the payload is a KUKA LBR iiwa 14 R820.

\begin{figure}
	\centering
	\includegraphics[width=8.5cm]{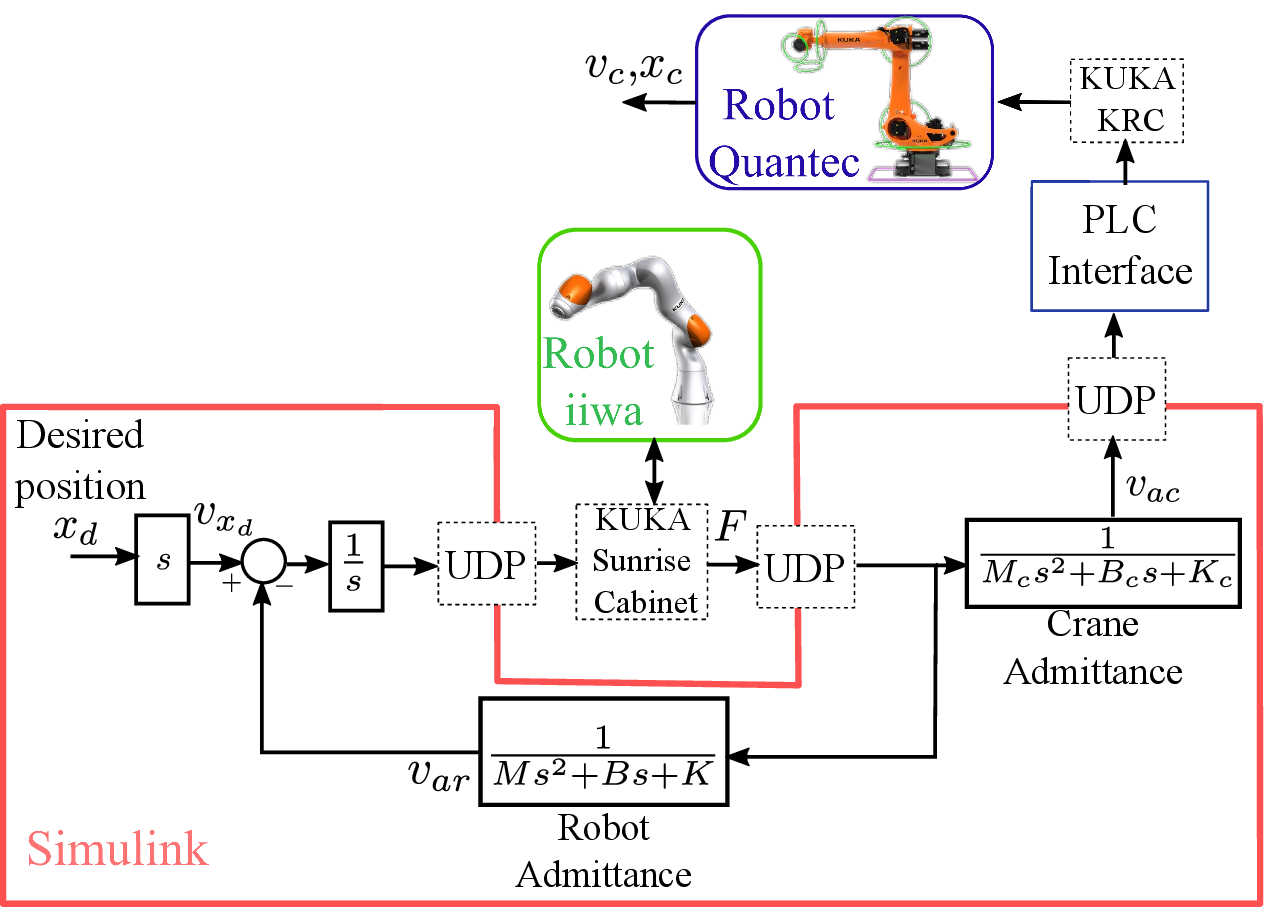}
	\caption{Implementation of the proposed scheme on the experimental setup}
	\label{fig:bdexpset}
\end{figure}

Fig. \ref{fig:bdexpset} shows a block diagram of collaborative scheme implementation. The crane is controlled with Cartesian velocity commands sent through a PLC interface. SW-in-the-Loop (SIL) approach was applied and a Matlab/Simulink interface \cite{Safeea2023} is used to send the robot's end-effector Cartesian position increments and receive the measured contact force $F$ using UDP protocol. The robot and crane admittance are implemented inside the Matlab/Simulink to send corrections to the robot and velocity commands $v_{ac}$ to the crane, respectively.

\begin{figure}
	\centering
	\includegraphics[width=9cm]{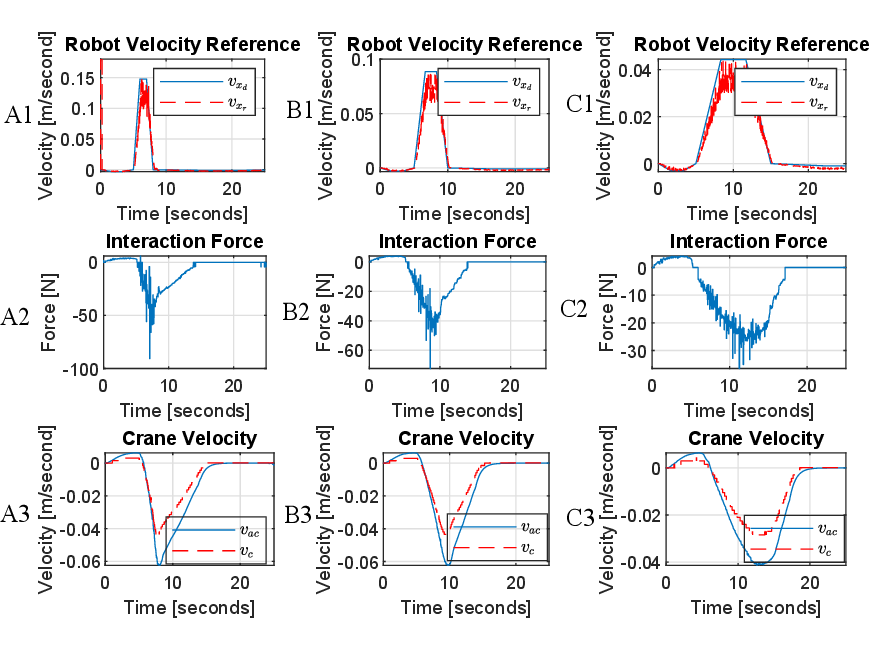}
	\caption{ Robot velocity reference $v_{x_d}$, Force $F$, and crane velocity command $v_{ac}$ and crane velocity $v_c$ from experimental results. Plots A1-A3, B1-B3, and C1-C3 correspond to the velocity commands 0.15~[m/s], 0.09~[m/s] and 0.045~[m/s], respectively.}
	\label{fig:Pexpfvc}
\end{figure}

From the mass  $m=10$~[kg] and length $R=0.5$~[m] of the pendulum attached to the industrial robot, the estimated stiffness is $K_e=200$~[N/m]. The admittance parameters for the robot and crane used in the experimental test are $M=10$~[kg], $B=1000$~[Ns/m], $K=2000$~[N/m], $M_c=1$~[kg], $B_c=500$~[Ns/m], and $K_c=600$~[N/m]. Three trapezoidal velocity profiles commanded to the robot are tested and the experimental results are presented in Fig. \ref{fig:Pexpfvc}. The maximum velocities of the velocity commands are 0.15~[m/s], 0.09~[m/s] and 0.045~[m/s]. The robot velocity reference $v_{x_d}$, the robot velocity $v_{x_r}$, the contact force $F$, the crane velocity command $v_{ac}$, and the velocity crane $v_c$ data was recorded and plotted in Fig. \ref{fig:Pexpfvc}. Plots A1-A3, B1-B3, and C1-C3 correspond to the velocity commands 0.15~[m/s], 0.09~[m/s] and 0.045~[m/s], respectively. From Fig. \ref{fig:Pexpfvc}, one can see that the robot and crane move together following the velocity profile and the velocity command generated by the interaction force $F$, respectively. 

\begin{figure}
	\centering
	\includegraphics[width=9cm]{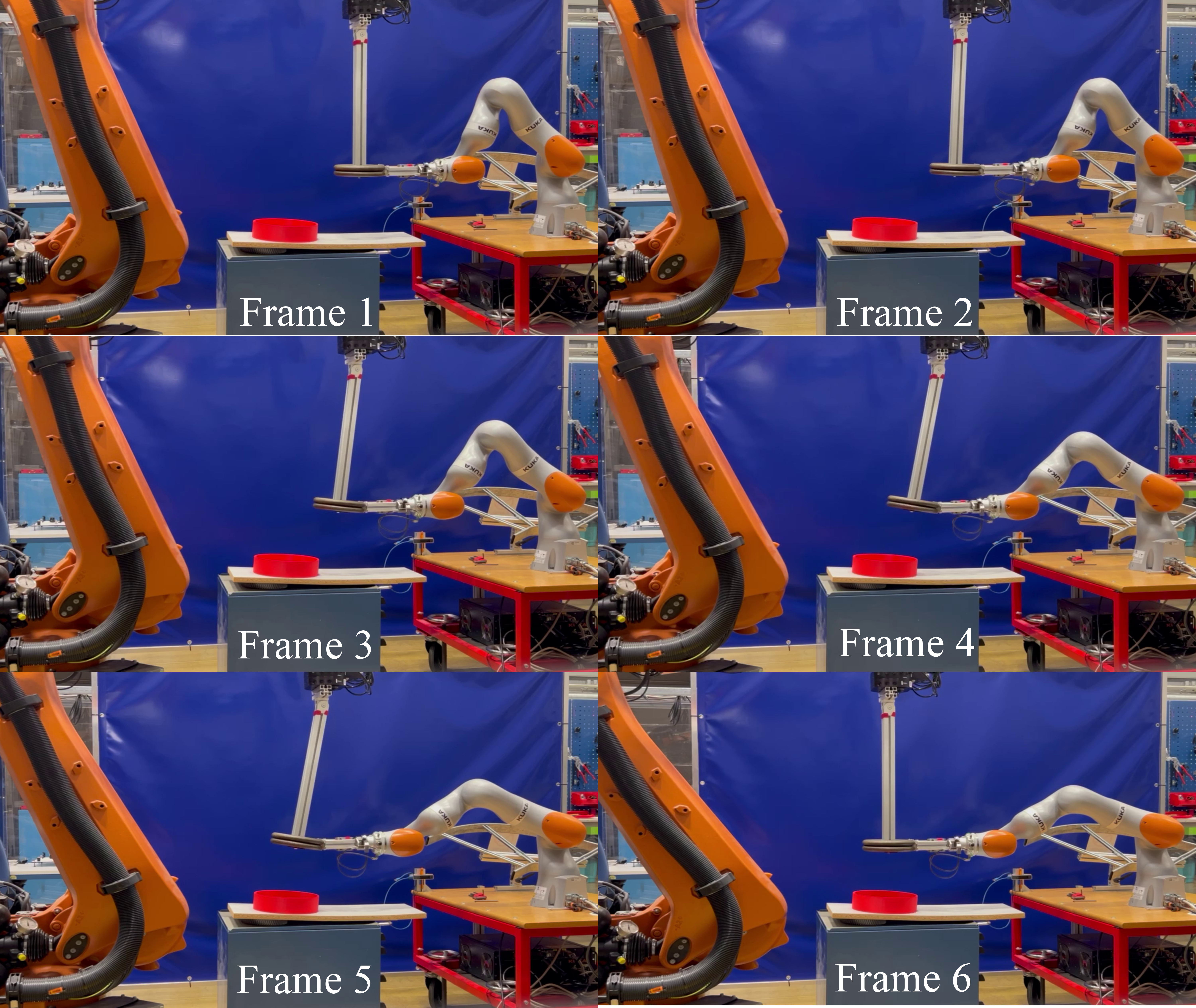}
	\caption{Experiment's frames of the robot-crane collaboration}
	\label{fig:framexp}
\end{figure}

Fig. \ref{fig:framexp} shows six video frames taken during the test corresponding to the 0.045~[m/s] profile. One can see the robot and crane moving collaboratively with the payload. 

\section{Conclusions}
This paper presents a collaborative control scheme for a robot and a crane that is useful for manipulating heavy payloads. The scheme is a safe approach since the operator is not in contact with the heavy object and compliance is considered in the robot and the crane. The simulations and experiments verified the functionality of the approach. Future work to be done is the extension from one-dimensional $x$ motion to three dimensions $xyz$. Also, eye-to-hand visual servoing will be integrated to get the payload's desired location and control the robot using visual feedback.

\nocite{*}
\bibliographystyle{agsm}
\bibliography{bio}
%
\vspace{1em} 
\noindent\textbf{Corresponding author}\\
Antonio Rosales can be contacted at: antonio.rosales@vtt.fi
%
\end{document}